\documentclass{article}

\usepackage[english]{babel}

\usepackage[letterpaper,top=2cm,bottom=2cm,left=3cm,right=3cm,marginparwidth=1.75cm]{geometry}

\usepackage{amsmath}
\usepackage{subfig}
\usepackage{graphicx}
\usepackage[colorlinks=true, allcolors=blue]{hyperref}

\title{MathGloss: Building mathematical glossaries from text}
\author{Lucy Horowitz \and Valeria de Paiva}

\begin{document}
\maketitle

\begin{abstract} 
 MathGloss is a project to create a knowledge graph (KG) for undergraduate mathematics from text, automatically, using modern natural language processing (NLP) tools and resources already available on the web.
 MathGloss is a linked database of undergraduate concepts in mathematics. So far, it combines five resources: (i) Wikidata,  
 a collaboratively edited, multilingual knowledge graph hosted by the Wikimedia Foundation, (ii) terms covered in mathematics courses at the University of Chicago, (iii) the syllabus of the French undergraduate mathematics curriculum which includes hyperlinks to the automated theorem prover Lean 4, (iv) MuLiMa, a multilingual dictionary of mathematics  curated by mathematicians, and (v) the nLab, a wiki for category theory also curated by mathematicians.
 MathGloss's goal is to bring together resources for learning mathematics and to allow every mathematician to tailor their learning to their own preferences. Moreover, by organizing different resources for learning undergraduate mathematics alongside those for learning formal mathematics, we hope to make it easier for mathematicians and  formal tools (theorem provers, computer algebra systems, etc) experts to “understand” each other and break down some of the barriers to formal math. 
\end{abstract}

\section{Introduction} 
When mathematicians read mathematical texts, the first thing they do is try to uncover the important concepts—what is the work about?
We can think of mathematics as an enormous web of definitions and concepts together with theorems that relate them. While it is hard for humans to learn how to produce \textit{proofs} of theorems, it should be much easier for anyone to find all the definitions required for the theorems and to organize them in structured ways.  Many  concepts and theorems are written down where almost anyone can find them, but some of it ``exists nearly as folklore" \cite{future}. That is, the math has been done, but it isn't organized in a way that is accessible to everyone outside the field. Ultimately, we hope to aid the organization of the math that is in this state by starting with the basics: an undergraduate curriculum.

Much of math accessible to a typical undergraduate is very well-understood and even well-organized in textbooks and online. However, all is not perfect. When asked to write down the definition of a “group,” different mathematicians may write down different things. For example, the authors of the Wikipedia article for “group”\footnote{\url{https://en.wikipedia.org/wiki/Group_(mathematics)}} ease the reader into abstraction, beginning with an explanation of the group structure of the integers and a quote about the mysterious nature of mathematical definitions before writing down the group axioms in careful detail. 
At a higher level, the authors of the nLab article for ``group"\footnote{\url{https://ncatlab.org/nlab/show/group}} jump right into ``monoid with inverses" and offhandedly reference some slightly nontrivial properties in the same sentence. A category theorist probably doesn't need a careful treatment of the group axioms, but someone learning group theory who happened to come across the nLab might be confused by this definition, while the Wikipedia article may have been more accessible. By putting them together, we can ensure common understanding across all levels of mathematics.

This principle extends to learning formal mathematics, which involves getting a computer to ``do" mathematics and verify all of the steps in a proof. No matter how abstract the nLab or Wikipedia entries may get, they were written for humans to read and understand. 
On the other hand, the definition of “group” in Lean, a formal theorem prover,\footnote{\url{https://lean-lang.org/}} is written for an audience of computers. 
Yet if we want people to learn to do formal mathematics (and we do, as it shows great promise for modern mathematics \cite{bigleage}), they will have to learn to interpret at least some things that were mostly meant for computers. 

Some people describe writing formal proofs as feeding definitions to a computer, and they note that the computers will often “whine” when they don’t “understand” something they’ve been given \cite{ai-is-coming}. If we can present formal mathematical concepts alongside the more familiar natural-language concepts, the communication between mathematician and computer will be greatly improved. Moreover, highlighting concepts that often appear in undergraduate curricula could help students taking and professors designing ``bilingual" math courses that teach both natural and formal proof-writing simultaneously.

This document serves as a written companion to a presentation given at the EuroProofNet joint meeting between the Workshop on Natural Formal Mathematics and the Workshop on Libraries of Formal Proofs and Natural Mathematical Language in Cambridge on September 7, 2023.\footnote{\url{https://europroofnet.github.io/cambridge-2023/\#horowitz}}


\section{Math concepts and Wikidata}
Wikidata \cite{wikidata2014} is a knowledge graph that contains the structured data behind Wikipedia. 
A knowledge graph represents a network of real-world entities—that is, objects, events, situations, or concepts—and the relationships between them. 
This information is usually stored in a graph database and visualized as a graph structure, hence the term knowledge “graph."

Wikipedia has a large number of articles on mathematical concepts, described in its ``Math Portal."\footnote{\url{https://en.wikipedia.org/wiki/Portal:Mathematics}}
We want to use WikiData as a knowledge repository for the math concepts we collect. 
Wikidata assigns a unique identification number of the form Qx...x (where x is a digit) to each  concept it describes. 
Some of the concepts are connected with descriptive links about their relationships to each other. For example, according to Wikidata a “book” (Q571)  is an “instance of” “written media.” 
These links are not present between every pair of items that “should” be linked because creating these links is a labor-intensive process. 

We attempt to map each term in our corpora to its Wikidata identifier (ID), and base our organization on common mappings.
That is, we present the terms that get mapped to the same Wikidata ID as the same mathematical concept.
Some of the mappings were done using \texttt{wikimapper},\footnote{\url{https://github.com/jcklie/wikimapper}} a Python package written by Jan-Kristoph Klie, a researcher at the Ubiquitous Knowledge Processing Lab of the Technical University of Darmstadt. 
Some of the mappings were done manually. 

The library \texttt{wikimapper} takes in the name of an “item” and produces a Wikidata ID whose page has a title matching that name. 
This often produces undesirable mappings due to the overloading of words in English and especially in mathematical English. 
For example, the words “group,” “ring,” and ``field” represent fundamentally important concepts in mathematics, but they also refer to everyday objects and can are often even used as verbs!
When “group” is put through \texttt{wikimapper}, the output is Q654302 instead of the hoped-for Q83478. 
The former is a “disambiguation page,” or a central hub page listing different things that go by the same name or similar names. 

We were able to almost completely resolve the disambiguation issue by appending parenthetical subject names to the ends of the terms. 
The parentheticals we used were “mathematics,” “linear algebra,” “algebraic geometry,” “calculus,” “category theory,” “commutative algebra,” “field theory,” “game theory,” “topology,” “differential geometry,” “graph theory,” “invariant theory,” “group theory,” “module theory,” “order theory,” “probability,” “statistics,” “ring theory,” “representation theory,” “set theory,” “string theory, “symplectic geometry,” and “tensor theory.” 
We chose these fields as parentheticals from the Wikipedia page listing the glossaries of mathematics.\footnote{\url{https://en.wikipedia.org/wiki/Category:Glossaries_of_mathematics}}
The name of the Wikipedia page describing mathematical groups is technically “Group (mathematics).” 
We were able to leverage this fact and the way that \texttt{wikimapper} can take a Wikipedia page name as input to drastically reduce the number of disambiguation pages returned. For the annotation of low resource domains, see \cite{klie-etal-2020-zero}, \texttt{wikimapper} is not perfect but helps considerably with the mapping task.

\section{Math Resources}

There are plenty of good sources of mathematics online. One of the issues beginners face is how to choose between these sources. We describe some of the sources we want to make available through MathGloss and we hope to be able to make other sources available in the near future.
Currently, MathGloss consists of terms collected from several resources for mathematical knowledge mapped to Wikidata, either manually or using \texttt{wikimapper}.

\subsection{Chicago} The Chicago corpus\footnote{\url{https://github.com/MathGloss/MathGloss/tree/main/chicago}} consists of approximately 700 terms related to courses in mathematics taken by the first author at the University of Chicago. 
With respect to MathGloss, it represents a “gold standard” of definitions of mathematical concepts that are well-known enough to appear in an undergraduate mathematics curriculum. That is, each entry in the corpus is annotated with its status as a definition. This corpus is not exhaustive, rather it reflects the first author’s interests and the topics covered within these interests by individual professors. Each concept in the corpus (e.g., ``group”) has its own Markdown file containing a definition of the term and links to the Markdown files corresponding to other terms. The links under the Chicago column on the MathGloss website lead to the content of these Markdown files, which contain links to other definitions.


\subsection{French undergraduate curriculum in Lean 4} This corpus consists of terms (translated from French) that are listed by the \textit{Ministére de l’Éducation Nationale et de la Jeunesse} as concepts undergraduate mathematics students are expected to know by the end of their degree. The Lean Community has added links from these concepts to their representation in Lean 4,\footnote{\url{https://leanprover-community.github.io/undergrad.html}} and the links on the MathGloss webpage lead to those Lean entries. Some terms do not have Lean counterparts yet, or the link to its counterpart is not included in the corpus. The mappings from terms to WikiData were done using \texttt{wikimapper}. There are 543 terms in total and 369 of them have Wikidata counterparts. Figure \ref{click} below shows such a mapping for the term ``0-1 Law".

\begin{figure}[h]
  \centering
  \subfloat{\includegraphics[width=0.45\textwidth]{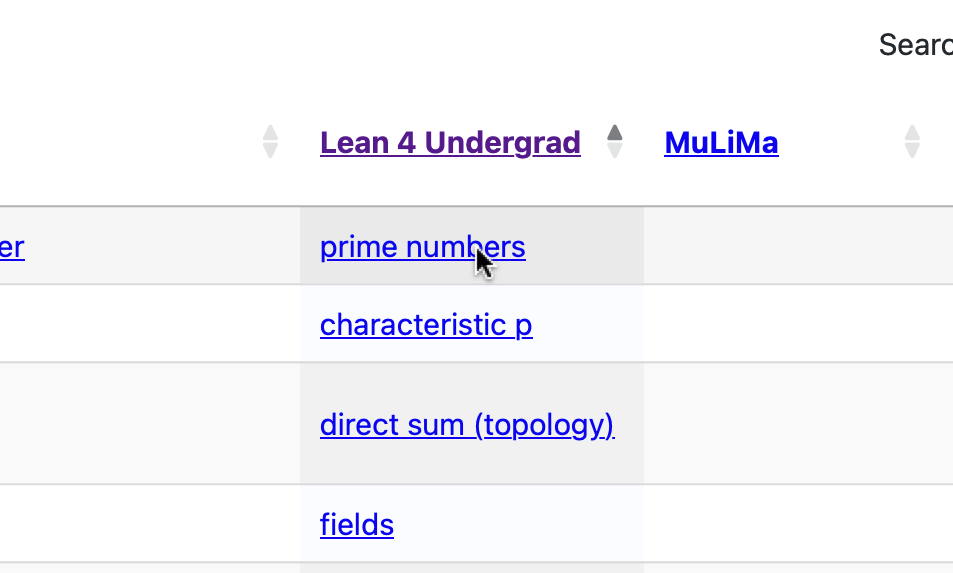}}
  \hfill\subfloat{\includegraphics[width=0.45\textwidth]{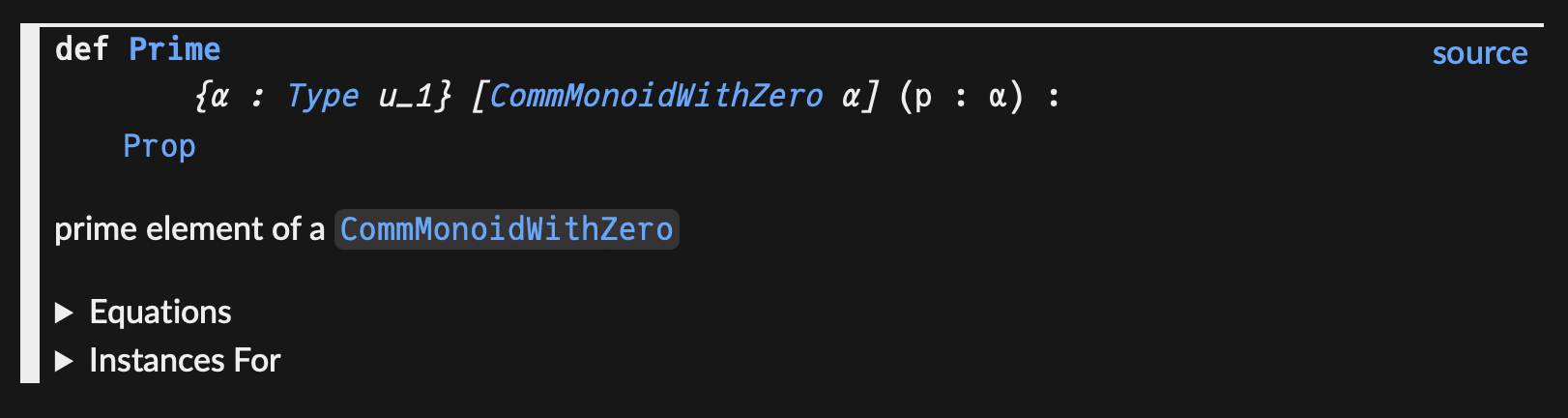}}
  \caption{Clicking through to Lean 4}\label{click}
\end{figure}

\subsection{Multilingual Mathematics (MuLiMa)}
The translation of mathematical terms between (natural) languages should be much easier than it actually is, given that terms in math are supposed to be unambiguous in their definitions. However, mathematicians are really at liberty to choose any name for any concept. Often, this means words that refer to the same mathematical object in two different languages will not ``translate" to each other. For example, the word in French for what we in English call a ``field" is ``corps," which literally means ``body." Moreover, much of mathematics that was first written about in English has no real translation into other languages—mathematicians will sometimes just use the English term. Collecting translations of mathematical concepts is therefore an important task. Tim Hosgood, a researcher at the Topos Institute, is working on this problem. He created a cross-language dictionary\footnote{https://thosgood.com/maths-dictionary/} for math with a similar structure to that of MathGloss. 
It has 305 terms at the moment, which were all manually mapped to Wikidata.

\subsection{nLab} The nLab\footnote{\url{https://ncatlab.org/nlab/show/HomePage}} is a wiki for higher mathematics, specifically category theory. It is not a resource intended for undergraduates, but we have included it here filtered along those terms which also appear in the other three corpora. 

The terms are the titles of the pages hosted on the nLab, but some of these are pages about people or books. The filtration by other corpora should ensure that only mathematical concepts make it into the final table. There were more than 18,000 page titles at the time of writing, and we found that 5377 of these had Wikidata items. Fewer than 5377 terms were included in the final table because of the filtering via the other resources.

\subsection{More online math} 
So far we have only included four corpora, which are not comprehensive of all undergraduate mathematics. One place we hope to look next is open-source textbooks for different topics and use natural language processing (NLP) to find important terms and concepts there.
Inspired by \cite{collard2022extracting}, we were successful in extracting terms from the journal Theory and Applications of Categories (TAC)\footnote{http://www.tac.mta.ca/tac/} using NLP before we decided to focus on undergraduate rather than research mathematics. 


\section{Using MathGloss}
\begin{figure}[h]
\centering
\includegraphics[width=0.9\textwidth]{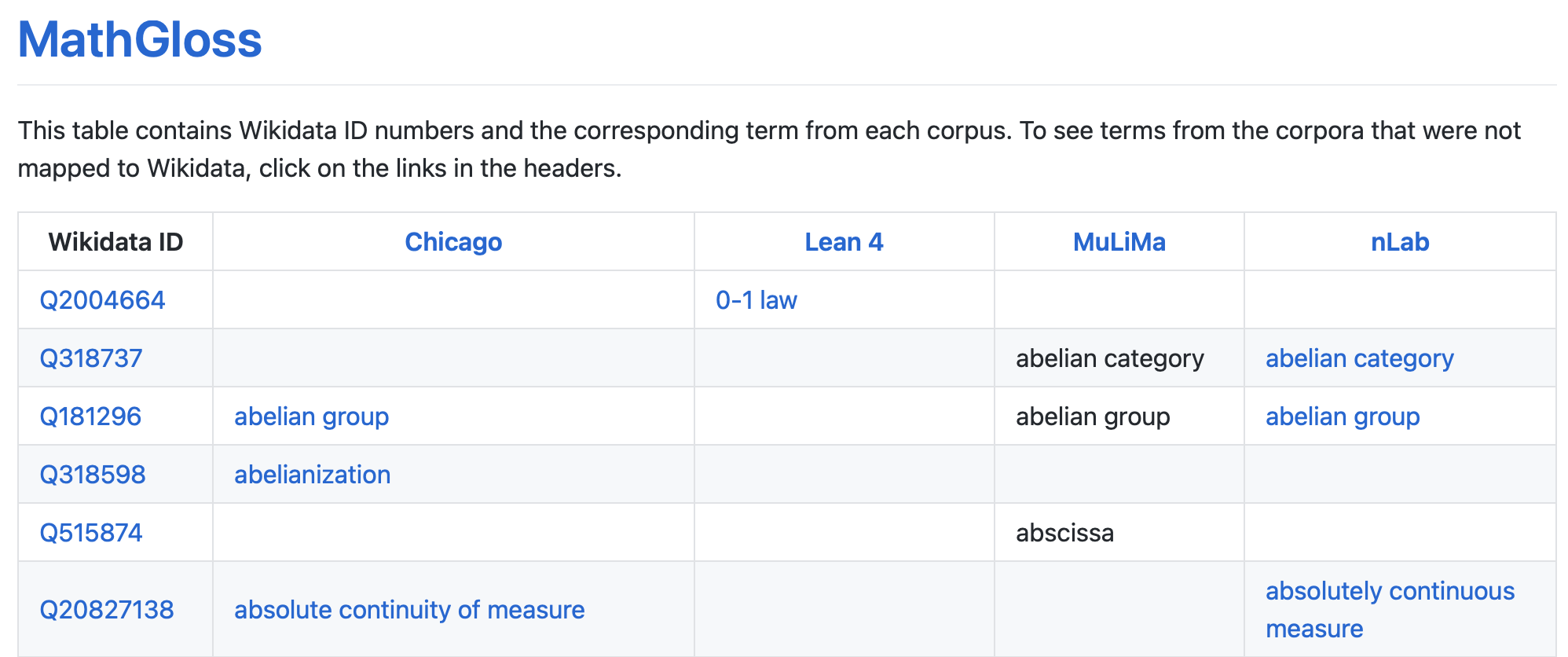}
\caption{The main table of MathGloss}\label{table}
\end{figure}

The table of terms in MathGloss can be found at \url{https://mathgloss.github.io/MathGloss/database}. Figure \ref{table} shows the first few rows of the table of mappings. As an example of how to use MathGloss, let's say we want to find out more about `abelian groups'. If we haven't already seen it in row three, a simple page search (Ctrl+F) can help find it in the table. 

Clicking on the link labeled ``Q181296" takes us to the Wikidata entry for abelian group, which contains much useful information about abelian groups and their relationships to other kinds of algebraic objects. In particular, it has a link to the Wikipedia page for ``Abelian group." Clicking on the link labeled ``abelian group" under the header ``Chicago" takes us to a definition of abelian group, hosted on the MathGloss website. This page contains links to other relevant definitions, also on MathGloss. There is currently no Lean 4 link for abelian group in the \textit{provided} list of undergraduate concepts, but if there were, clicking on it would take us to the entry in the Lean 4 documentation defining abelian groups. Since ``abelian group" appears under the MuLiMa heading, going to the MuLiMa website will allow us to see its translation into several languages. Finally, if we click on the link under the nLab heading, we will see the nLab page for ``abelian group," which takes on a distinctly categorical point of view. 

The lack of a link to an instance of ``abelian group" in Lean 4 highlights the need to collect information from multiple resources. Certainly one can talk about abelian groups in Lean, but the list of undergraduate concepts we used just happens not to link there. 

\section{Tools for NLP}\label{nlp}
At another stage in the project, we collected terms from the abstracts of articles published in \textit{Theory and Applications of Categories} (TAC) using natural language processing (TAC). We did not include it in this iteration of MathGloss because as a corpus of research mathematics, it does not fit our goal of organizing undergraduate math. However, we hope to apply this technique to other corpora of undergraduate mathematics in the future. We describe the technique below.

To extract terms from TAC, we used the Python library spaCy\footnote{\url{https://spacy.io/}} to perform grammatical analysis on the text of the 755 abstracts from the articles in TAC.\footnote{\url{https://github.com/ToposInstitute/tac-corpus}}
SpaCy performs syntactic parsing of sentences using the Universal Dependencies (UD)\cite{nivre-etal-2016} framework. UD is an open-source project that works towards standardizing grammatical annotation to make linguistics research more consistent. The output of this analysis is in a UD-developed format called CoNLL-U,\footnote{\url{https://universaldependencies.org/format.html}} which we then inspect using a script from UD.
 
\begin{figure}[h]
\centering
\includegraphics[width=0.9\textwidth]{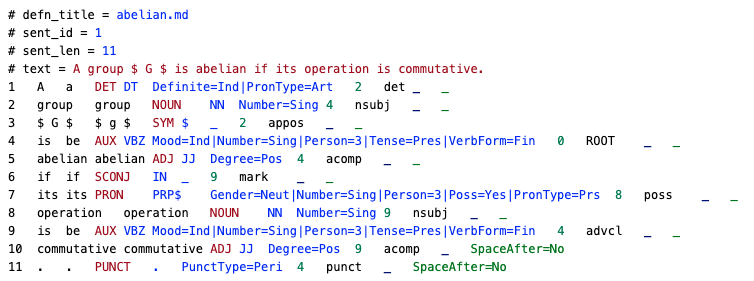}
\caption{Analysis of the definition of ``abelian group" in  CoNLL-U format}\label{conll}
\end{figure}

 A CoNLL-U file is a plaintext file that displays sentence analysis in a particular structure. It allows comment lines, which are indicated by “\texttt{\#}”, and in the CoNLL-U files we created, the comments include the text of the sentence, the number of its entry in the corpus, and the length (in “tokens”) of the sentence. Figure \ref{conll} shows an example of a CoNLL-U file containing a sentence analyzed in this way after application of the ``detextor" pipeline component described below. Each word in the sentence is written on its own line following the text of the sentence along with information about the word, for example its part of speech.

At its simplest, spaCy takes in a section of text (called a Doc, short for ``document”), and then uses its models to split the text into sentences, split sentences into tokens, and then assigns parts of speech and assigns Universal Dependency relations to each token. A token can be thought of as a generalization of a word: it can be a punctuation mark, or especially in our case, a piece of mathematical notation. The process of splitting a Doc into sentences is called ``sentencization,” and the process of splitting a sentence into tokens is called ``tokenization.” We used the smallest model provided by spaCy, called ``\texttt{en\_core\_web\_sm},” because of our limitations in computing power. Using larger models made no difference in output, but took more time and resources to generate that output.

It is possible to run spaCy on \LaTeX{} code, but without making specific modifications for mathematical notation, the results are very poor due to incorrect sentencization (including both sentences that are too long and sentences that are too short) and the over-tokenization of the sentence. \LaTeX{} code consists of many punctuation marks and commands that often resemble English words, but none of these things should be parsed as normal English. Without modifications to the default spaCy pipeline, each piece of the code that represents a mathematical expression is fragmented and treated separately. Our goal is to extract definitions and concepts from texts written in mathematical English, and \LaTeX{} is an integral part of that.

First-year mathematics students are taught that mathematical expressions should always be situated within a grammatically correct sentence. They should never stand on their own, and the statement ``$x = y$” should be read ``$x$ equals $y$,” and can therefore be considered an independent clause. Within abstracts and definitions, one does not usually make declarative statements like “$x = y$,” so most of the \LaTeX{} we encounter in our corpora should be thought of as nouns or as names of instances of mathematical objects. Because code itself is not “natural language,” one way to analyze it is to take each piece of \LaTeX{} code to represent a single ``word."

This makes up our main approach to solving the tokenization problem. It is as simple as telling spaCy to treat everything in between two dollar signs as one token. The implementation of this pipeline component is done using the ``retokenizer” context manager and requires that dollar signs (and for best results, hyphens) be padded with spaces. This new pipeline component, called ``detextor,” is implemented after the ``tagger” component, which assigns parts of speech to tokens. The detextor pipeline component represents only a partial solution to the problem of annotating \LaTeX{} code as it struggles to capture the actual information conveyed by formulas. However, it greatly improves the accuracy of sentencization, which is crucial for annotating regular English words with their parts of speech. 

The part-of-speech annotation forms the foundation of our term extraction. We are able to pick out what we expect to be mathematical terms or concepts with a simple heuristic. From the CoNLL-U files, we are able to compile lists of ``lemmas," or basic forms of words (e.g. ``be" for the word ``is," or ``group" for the word ``groups") according to their part of speech. We suppose that the most frequently occurring nouns, adjective-noun phrases, and ``compounds" are math terms. ``Compound" is a UD annotation for certain multi-word phrases that represent a single thing\footnote{\url{https://universaldependencies.org/u/dep/compound.html}}. Adjective-noun phrases consist of any number of adjectives followed by a noun or by a compound. The upper ends of the frequency tables for these types of terms are almost exclusively populated by math concepts, but we do not yet know if the lists exclude some concepts. 

This process is what we hope to use on other bodies of mathematical text when they do not themselves provide ready-made lists of terms. Unfortunately, it does not produce any kind of link to an explanation or definition of a given term. In the future we hope to be able to pinpoint the definitions of these extracted terms and give pointers to their locations in text. In other words, we want to perform entity linking on math text.

\section{Future Work}
First, we hope to develop a method to more easily collect and map terms from different math resources. In particular, we want to include more theorem provers in our mappings. Presently we are looking into adding links to Agda, Coq, and Isabelle. At the Dagstuhl seminar on automated mathematics,\footnote{\url{https://www.dagstuhl.de/23401}} there was discussion on how best to find and compile instances of undergraduate math concepts from 
these provers. Hopefully they will follow the example of Lean and produce such resources.

On the natural language side of collecting more terms from more resources, we want to use machine learning techniques to extract definitions of the terms we already know how to find. Some work has been done in this direction, even specifically tailored to mathematical text \cite{vanetik-etal-2020-automated}, but results do not look impressive \cite{collard2023}.

Another future project is to come up with a way to verify that the mappings to Wikidata are indeed correct. The addition of subject parentheticals definitely reduces the number of disambiguation pages output by Wikimapper, but some still slip through the cracks. We could potentially do this by verifying that the terms we map to a Wikidata item have the same relations to other Wikidata items—this relies on the linking that already exists within corpora like Chicago Notes.

Currently, the process for performing mappings is labor-intensive even though it is automated to some degree. We hope to further automate the process to take advantage of the fluid nature of Wikidata, which is always adding more terms. Relatedly, there are some basic improvements that need to be made to the GitHub Pages website. As we add new resources, we want users to be able to select which ones they see on the webpage at any time and for there to be a better search function than pressing  Ctrl+F. Moreover, the website needs some support for \LaTeX{} in order to properly display the content in Chicago Notes that is also hosted there.

\section{Conclusion}
MathGloss aims to help people from different backgrounds make sense of the diverse resources for mathematics available online through organization by individual concept. 
Should one encyclopedia’s article for a particular construction be inscrutable, we would like it to be easy to find another article that is more closely aligned with the reader’s background and therefore easier to understand. 
Moreover, MathGloss represents a step in the direction of bridging the gap between natural math as done by humans and formal math. By creating a knowledge graph of undergraduate mathematics, we hope to empower students, mathematicians, and those who use mathematics in their work to both better navigate the intricate web of definitions and theorems and to embrace the use of formal systems. 

\bibliographystyle{alpha}
\bibliography{references}
\end{document}